\documentclass{article}
\usepackage{spconf,amsmath,graphicx}

\usepackage{enumitem}
\setlist{nosep, leftmargin=14pt}
\usepackage{booktabs,array}
\usepackage{tabularx}
\usepackage{float}
\usepackage{ragged2e}
\usepackage{array}
\usepackage[table]{xcolor} 
\usepackage{amsfonts}
\newcolumntype{Y}{>{\RaggedRight\arraybackslash}X}
\usepackage{mwe} 
\usepackage{amsmath,mathtools}
\usepackage{comment}
\newcommand{\LN}{\mathrm{LN}}
\newcommand{\MLP}{\mathrm{MLP}}
\newcommand{\MHA}{\mathrm{MHA}}
\newcommand{\softmax}{\operatorname{softmax}}

\title{DM-QPMNet: Dual-Modality Fusion Network for Cell Segmentation 
in Quantitative Phase Microscopy}
\name{Rajatsubhra Chakraborty$^{1}$,  Ana Espinosa--Momox$^{2}$,  Riley Haskin$^{2}$,  Depeng Xu$^{1,3}$,  Rosario Porras--Aguilar$^{2,3}$}

\address{$^{1}$College of Computing and Informatics, University of North Carolina at Charlotte, NC, USA\\
$^{2}$Department of Physics and Optical Science, University of North Carolina at Charlotte, NC, USA\\
$^{3}$Center for TAIMing AI, University of North Carolina at Charlotte, NC, USA}

\begin{document}
%
\maketitle
 \begin{abstract}
Cell segmentation in single-shot quantitative phase microscopy (ssQPM) faces challenges from traditional thresholding methods that are sensitive to noise and cell density, while deep learning approaches using simple channel concatenation fail to exploit the complementary nature of polarized intensity images and phase maps. We introduce DM-QPMNet, a dual-encoder network that treats these as distinct modalities with separate encoding streams. Our architecture fuses modality-specific features at intermediate depth via multi-head attention, enabling polarized edge and texture representations to selectively integrate complementary phase information. This content-aware fusion preserves training stability while adding principled multi-modal integration through dual-source skip connections and per-modality normalization at minimal overhead. Our approach demonstrates substantial improvements over monolithic concatenation and single-modality baselines, showing that modality-specific encoding with learnable fusion effectively exploits ssQPM's simultaneous capture of complementary illumination and phase cues for robust cell segmentation.
\end{abstract}


\section{Introduction}
Deep-learning cell segmentation has advanced rapidly through large, labeled fluorescence datasets paired with standardized training pipelines. Generalist models such as Cellpose learn robust appearance priors from heterogeneous corpora and transfer across laboratories and cell types~\cite{Stringer2021Cellpose}, while self-configuring frameworks like nnU-Net codify best practices for preprocessing, architecture scaling, and training~\cite{Isensee2021nnUNet}. Promptable foundation models such as SAM have further extended segmentation to novel distributions via interactive prompts~\cite{Kirillov2023SAM}. These systems demonstrate impressive breadth but require abundant labeled images with conspicuous boundaries, typically achieved through fluorescence staining.

Fluorescence microscopy is powerful yet operationally expensive and scientifically limiting. Assembling large, expertly annotated datasets (e.g., LIVECell) demands sustained curation effort~\cite{Edlund2021LIVECell}, and models often exhibit brittleness across different stains, instruments, and protocols~\cite{Ounkomol2018LabelFree}. Beyond annotation costs, fluorescence introduces photobleaching and phototoxicity that constrain live-cell imaging and longitudinal assays~\cite{Laissue2017Phototoxicity}, limiting throughput for routine cell analysis. \textbf{Quantitative phase microscopy (QPM)} offers a label-free alternative by measuring optical path length to reveal cell mass and morphology without exogenous contrast agents~\cite{Park2018QPIReview}. However, conventional QPM methods face practical limitations: phase-shifting techniques (DHM, DPM)~\cite{Kim2010DHMReview, Kemper2008DHM} require temporal multiplexing or complex interferometric alignment, while transport-of-intensity approaches~\cite{Zuo2015TIE} necessitate mechanical scanning. These constraints increase system cost and reduce robustness. \textbf{Single-shot quantitative phase microscopy (ssQPM)} addresses these limitations by employing polarization-sensitive cameras that simultaneously record four interferograms at distinct polarization angles (0°, 45°, 90°, 135°) and reconstruct a pixel-aligned quantitative phase map in one exposure~\cite{EspinosaMomox2024BOE}. This common-path architecture operates under low-intensity illumination, exhibits inherent vibration robustness, and enables real-time acquisition at camera-limited frame rates~\cite{Millerd2017InstantaneousPhase}. Critically, ssQPM provides \textit{complementary multi-modal data}: polarized intensities encode high-frequency edge and texture features, while the phase map captures low-frequency optical thickness and mass distribution both acquired simultaneously in a single shot.

Traditional QPM segmentation relies on intensity-driven thresholding and morphological operations~\cite{EspinosaMomox2024BOE}, which exhibit sensitivity to speckle noise and cell density~\cite{Debeir2008PhaseWatershed}. Learning-based approaches can overcome these limitations~\cite{Park2018QPIReview}, but most prior work treats QPM as a \textit{single-modality} problem segmenting from phase maps alone or individual polarization angles. In medical imaging, \textbf{multi-modal fusion architectures} have consistently demonstrated that combining complementary channels improves segmentation accuracy~\cite{Zhou2020UNetSurvey}. Early fusion via input concatenation is straightforward but can dilute modality-specific statistics when channels exhibit different frequency characteristics or dynamic ranges~\cite{Dolz2019DenseMultiPath}. Dual-encoder designs that learn modality-specific representations before interaction have proven effective for multi-sequence MRI and multi-stain histology~\cite{Oktay2018AttentionUNet,  Valindria2018HeteroModal}, with attention mechanisms enabling content-aware fusion at intermediate scales~\cite{Chen2021TransUNet, Valanarasu2021MedT}. Frameworks like nnU-Net provide stable decoder topologies and training heuristics~\cite{Isensee2021nnUNet}, offering a robust foundation for architectural extensions.

\textbf{This work addresses single-shot QPM segmentation and asks:}
\textit{\textcolor{blue}{Can we effectively leverage complementary multi-illumination intensity and phase cues for accurate cell segmentation?}}
Two naive baselines exist: (1) \textit{single-modality} approaches that segment from phase maps alone or concatenated polarization angles only, and (2) \textit{early-fusion multi-modal} approaches that concatenate all five channels (four polarization angles plus phase) into a standard nnU-Net. Single-modality methods discard either the high-frequency edge information (phase-only) or low-frequency thickness cues (angles-only), limiting segmentation accuracy. Early fusion, while incorporating both modalities, forces complementary signals through shared normalization and convolutional stems, potentially diluting modality-specific contributions. Polarized angles capture high-frequency boundary information while the phase map encodes low-frequency thickness distributions treating these as a homogeneous input overlooks their distinct statistical properties and spatial frequency content.

We posit a \textbf{dual-encoder architecture with late fusion at intermediate depth} that respects modality-specific characteristics. Our model separates the four polarized intensity images and the quantitative phase map into distinct encoders, allowing each stream to develop appropriate feature representations independently. At an intermediate encoder depth, where receptive fields span cell-scale context while retaining spatial detail, we apply \textbf{multi-head attention (MHA)} to fuse features in a content-aware manner. This enables polarized edge and texture representations to selectively query the phase encoder for complementary optical thickness information. Following fusion, a shared encoder tail and standard nnU-Net decoder with deep supervision complete the segmentation pipeline. The design preserves nnU-Net's training stability and decoder topology while introducing principled multi-modal integration at minimal parameter overhead.

\noindent\textbf{Our contributions are:}
\begin{itemize}[leftmargin=*, itemsep=2pt]
    \item A dual-encoder nnU-Net with mid-encoder fusion via multi-head attention that treats polarized intensity and phase as distinct modalities;
    \item Systematic ablation studies comparing fusion strategies (early vs. late, attention vs. concatenation), interaction depths, and modality contributions (single-modality vs. multi-modal);
    \item Architecture-level framework demonstrating that multi-modal fusion with modality-specific encoding outperforms both single-modality and early-fusion baselines for ssQPM cell segmentation.
\end{itemize}

\section{Methods}
\label{sec:methodology}

\subsection{Dataset}
\begin{figure}
    \centering
    \includegraphics[width=1\linewidth]{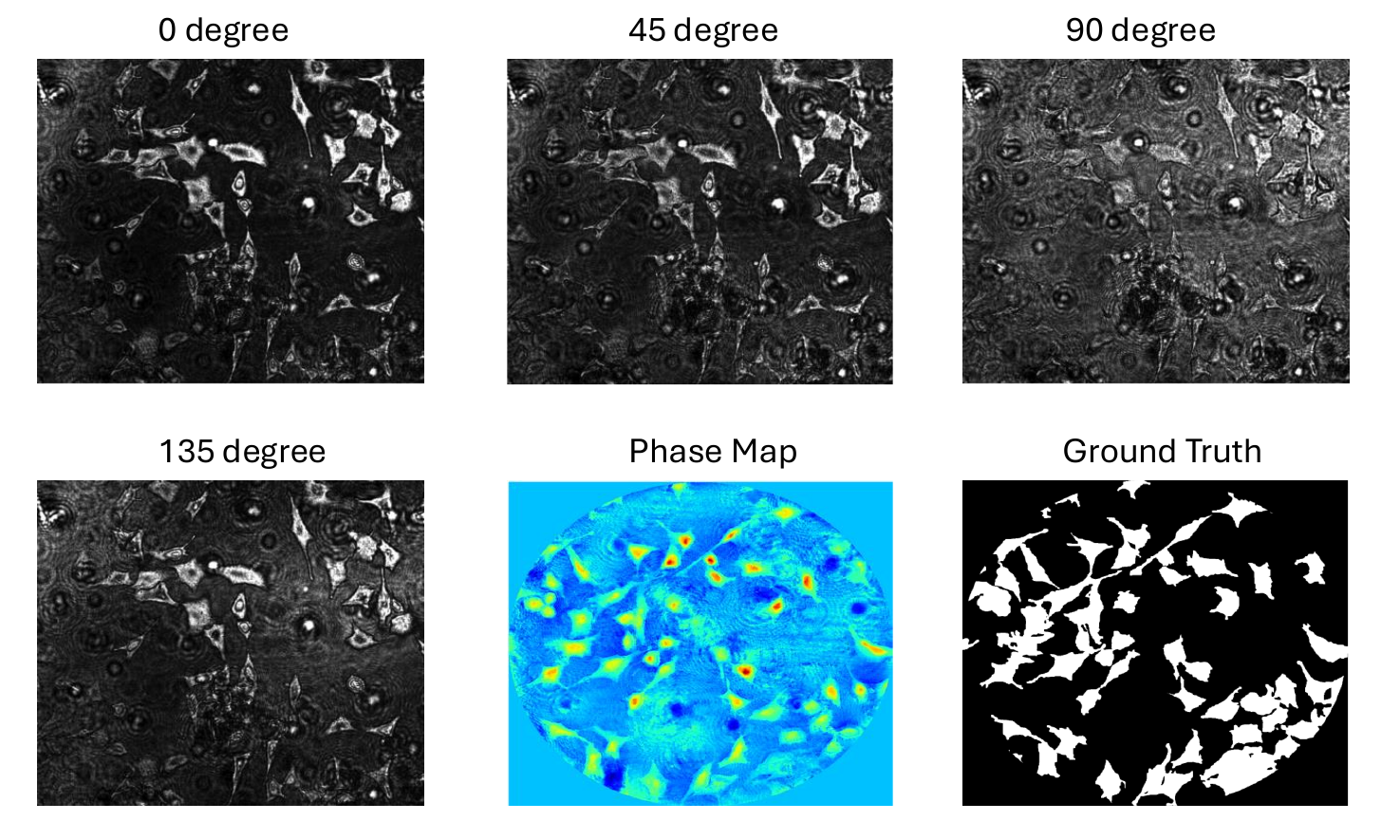}
    \caption{ssQPM dataset inputs: four polarized intensity images (0°, 45°, 90°, 135°), one quantitative phase map, and the corresponding ground-truth binary cell mask.}
    \label{fig:dataset}
\end{figure}

Following the data collection and processing in \cite{EspinosaMomox2024BOE}, we obtain a single\textendash shot quantitative phase microscopy (ssQPM) dataset acquired with a polarization\textendash sensing camera that simultaneously records four interferograms at 0\textdegree, 45\textdegree, 90\textdegree, and 135\textdegree as seen in Fig.\ref{fig:dataset} and reconstructs a quantitative phase map in the same exposure. This common path system operates under low intensity illumination, is robust to vibrations, and is positioned as a cost effective alternative for routine cell analysis.It couples intensity maps from 4 polarization components with an aligned phase map in a single shot.We posit that these complementary cues enable more reliable cell segmentation than any single modality alone (angles or phase).In our experiments, we use 24 samples for training and 6 samples for testing, spanning high, medium, and low confluence regimes.

\subsection{Input Representation and Modality Handling}
We process the images obtained from ssQPM into four polarized intensity images and one quantitative phase map. Let
$\mathbf{A}\!\in\!\mathbb{R}^{H\times W\times 4}$ denote the stacked angles
$\{0^\circ,45^\circ,90^\circ,135^\circ\}$ and
$\mathbf{P}\!\in\!\mathbb{R}^{H\times W\times 1}$ the phase map, pixel–aligned with $\mathbf{A}$.
To respect modality statistics, we apply \emph{per–modality} normalization:
each angle channel is normalized by its own $(\mu,\sigma)$ estimated over the training set, and the phase is normalized by robust quantiles (clipped to the $[q_{0.5\%},q_{99.5\%}]$ range and rescaled). The input to the network is \emph{not} concatenated; instead, angles and phase are routed to distinct encoders (\S\ref{sec:arch}), allowing modality–specific filters to emerge prior to fusion.

\subsection{Architecture Overview}
\label{sec:arch}
\begin{figure}[!t]
  \centering
  \includegraphics[width=0.83\columnwidth]{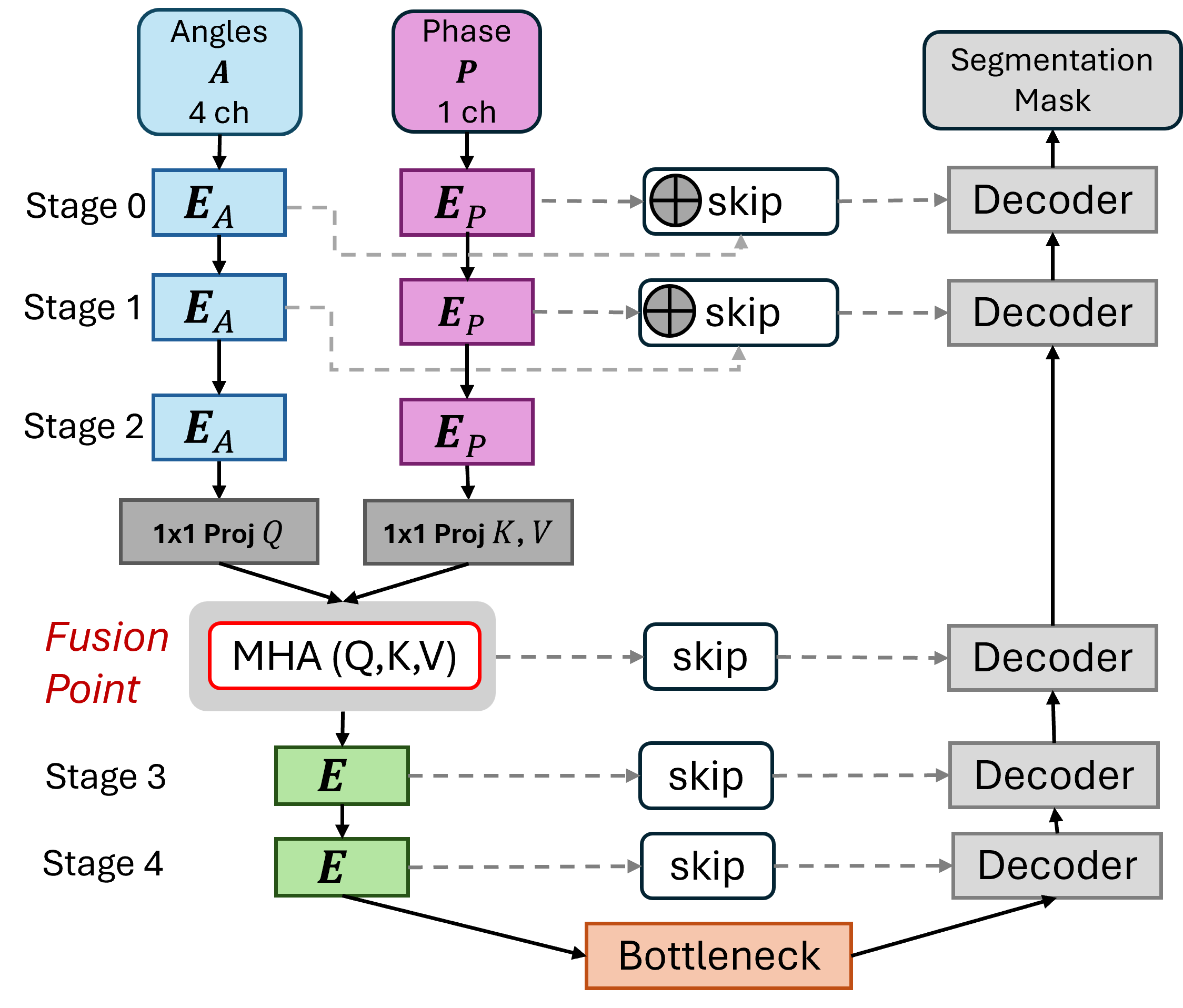}
  \caption{Late-fusion dual-encoder nnU-Net: four polarized intensity inputs (0°, 45°, 90°, 135°) and one phase map enter separate encoders; features fuse at Stage~2 via multi-head attention, then pass through a shared encoder tail and a decoder with deep supervision to produce the cell mask.}
  \label{fig:arch}
\end{figure}

The overall architecure of our model is shown in Figure \ref{fig:arch}. Our design uses a 2D nnU\textendash Net macro–architecture (five encoder stages, symmetric decoder with deep supervision) as backbone, but introduces three targeted changes for the dual modality learning: (i) \textbf{dual encoders} for angles and phase, (ii) \textbf{late fusion} at an intermediate depth via multi–head attention (MHA), and (iii) \textbf{dual–source skip aggregation} in the decoder.

\subsubsection{Dual encoders with late MHA fusion.}
Two parallel encoders $\mathbf{E}_A$ (angles) and $\mathbf{E}_P$ (phase) process $\mathbf{A}$ and $\mathbf{P}$, respectively. Each of the five encoder stages (0 to 4) consists of strided downsampling (stride $2$) followed by $n$ nnU\textendash Net blocks (Conv$3\!\times\!3$ \,$\to$\, Norm \,$\to$\, LeakyReLU).
We retain stage–wise feature maps as skip candidates.
At Encoder Stage 2 (spatial scale $H/4\times W/4$), we align channel dimensions with $1{\times}1$ projections and  then apply multi–head attention to \emph{fuse} features with angles as queries $Q$ and phase as key $K$ and value $V$. 
It is followed by a residual + MLP post–fusion block, where MLP is a $1{\times}1$ conv bottleneck with nonlinearity.
This \emph{directed} attention (angles query the phase) is chosen to let high–frequency edge/texture cues from polarized views pull in complementary thickness cues from phase. (A symmetric variant can be realized by an additional MHA term with roles swapped; we keep the single–direction scheme for efficiency.)
From Encoder Stage 3 onward, we use a \emph{single} encoder operating on the fusion block, where Stage 4 is followed by the bottleneck block (no skip). 

\subsubsection{Decoder with dual–source skip aggregation.}
For each of the five decoder Stages (4 to 0), we upsample and concatenate with a \emph{fused} skip constructed from both encoders at the corresponding scale. Pre–fusion (stages $0,1$), we produce skip tensors via concatenating  skip residuals from angles and phase from corresponding encoder stages followed by aggregation blocks (Conv$1\!\times\!1$ \,$\to$\, Norm \,$\to$\, Act).
It compresses and balances contributions from both modalities before passing to the decoder.
Post–fusion (stages $3,4$), we use skip residuals from corresponding fusion blocks as standard skips.
The decoder applies transposed–conv upsampling, concatenation with skip tensors, and nnU\textendash Net blocks to refine predictions. Deep supervision heads are attached at intermediate decoder resolutions as in nnU\textendash Net.

Relative to a monolithic 5–channel nnU\textendash Net (all inputs concatenated at the image level), our design introduces:

\begin{enumerate}[leftmargin=12pt,itemsep=2pt,topsep=2pt]
\item \textbf{Modality\textendash specific stems and encoders:}
   Separate early filters for angles vs.\ phase prevent the phase channel’s low–frequency structure from being swamped by the higher–variance polarized intensities, and allow different normalization/activation statistics to stabilize training.

\item \textbf{Attention\textendash based late fusion:}
   Instead of fixed concatenation, MHA learns a \emph{content–aware} mixing of phase and angle features at the mid–encoder resolution (Stage~2), where receptive fields are large enough to resolve cell–scale context but retain spatial detail.

\item \textbf{Dual–source skip fusers:}
   Skip tensors from the two encoders are compressed by a learned $1{\times}1$ mixer before concatenation in the decoder, reducing channel imbalance and avoiding domination by any single modality.

\item \textbf{Projection alignment and residual post\textendash fuse block:}
   $1{\times}1$ projections align feature widths prior to attention; residual MLP stabilization after attention preserves gradient flow (Transformer–style pre–norm), which nnU\textendash Net does not require in its single–stream design.

\item \textbf{Per–modality normalization and clipping:}
   Angle channels and phase are normalized with modality–appropriate statistics, improving conditioning compared to a single affine normalization over concatenated inputs.
\end{enumerate}

\section{Results}
\subsection{Implementation Details}
All models are trained using the 2D nnU-Net pipeline with its default preprocessing, patch-based sampling, data augmentation, deep supervision, and optimizer schedule. We report mean Dice and Intersection-over-Union (IoU) along with standard deviation across the 6 test samples. All nnU-Net style baselines (5-channel early fusion, angles-only, and phase-only) and DM-QPMNet are trained for 300 epochs under identical settings. We also fine-tune Cellpose--SAM on the phase map alone for comparison. 
\subsection{Result Discussion}
Table~\ref{tab:quant} reports segmentation performance across six held-out test samples. Our dual-encoder architecture (DM-QPMNet) achieves Dice $0.888\pm0.026$ and IoU $0.799\pm0.040$, consistently surpassing the monolithic 5-channel baseline (Dice $0.863\pm0.026$, IoU $0.782\pm0.041$). Although the absolute improvement is modest ($+0.025$ Dice, $+0.017$ IoU), the gain is coherent across metrics under identical training protocols, confirming that separate encoders with attention-based fusion provide reproducible benefits over naive concatenation. Multi-illumination models clearly dominate single-modality variants: angles-only trails by $-0.031$ Dice and $-0.043$ IoU, while phase-only severely underperforms (Dice $0.472\pm0.246$). Cellpose-SAM trained on phase recovers substantial ground versus phase-only U-Net ($+0.258$ Dice) but remains well below multi-illumination configurations ($-0.158$ Dice vs. ours). Notably, multi-illumination models exhibit low variance ($0.026$), whereas angles-only shows markedly higher variability ($0.084$), indicating greater sensitivity to scene content when phase information is absent.

\begin{table}[hb]
\centering
\small
\caption{Segmentation performance on QPM test images.}
\label{tab:quant}
\resizebox{\columnwidth}{!}{%
\begin{tabular}{lcc}
\hline
\textbf{Model} & \textbf{Dice (mean $\pm$ std)(↑)} & \textbf{IoU (mean $\pm$ std)(↑)} \\
\hline
\rowcolor{gray!15}
\textbf{DM-QPMNet (ours)}              & \textbf{0.888 $\pm$ 0.026} & \textbf{0.799 $\pm$ 0.040} \\
nnU-Net (5-channel: 0\textdegree,45\textdegree,90\textdegree,135\textdegree, phase) & 0.863 $\pm$ 0.026 & 0.782 $\pm$ 0.041 \\
nnU-Net (4-channel: angles only)                 & 0.857 $\pm$ 0.084 & 0.756 $\pm$ 0.116 \\
Cellpose-SAM (phase only, fine-tuned) & 0.730 $\pm$ 0.209 & 0.604 $\pm$ 0.213 \\
nnU-Net (phase only)                              & 0.472 $\pm$ 0.246 & 0.336 $\pm$ 0.204 \\
\hline
\end{tabular}%
}
\end{table}

Table~\ref{tab:fusion_ablation} isolates fusion design elements by varying operator type and interaction depth. With fusion at Stage 2, replacing multi-head attention with concatenation plus $1{\times}1$ projection reduces Dice/IoU by approximately 1\%. Cross-gating performs 1.5\% worse, while early fusion (5-channel input concatenation) lags by 1.7\%. Varying fusion depth reveals a clear optimum at Stage 2: Stage 1 fusion incurs a 1.9\% penalty, while Stage 3 trails by 1.2\%. These results confirm that content-aware fusion at intermediate encoder depth maximizes performance. 

\begin{figure}[!t]
  \centering
  \includegraphics[width=0.8\columnwidth]{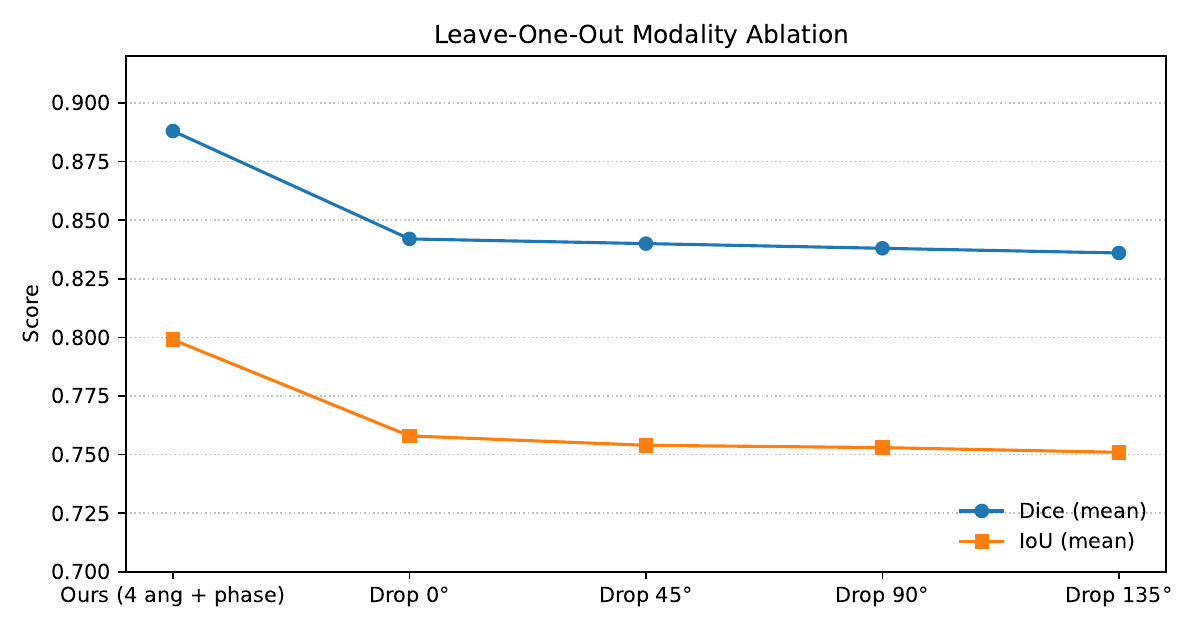}
  \caption{Leave-one-out modality ablation showing mean Dice and IoU over 6 test samples.}
  \label{fig:loo_ablation_nophase_line}
\end{figure}

Figure~\ref{fig:loo_ablation_nophase_line} presents leave-one-out analysis across the four polarization angles. Removing any single angle produces small, near-symmetric declines in Dice and IoU, with tightly clustered curves indicating no uniquely critical orientation. This demonstrates that robustness stems from angular diversity rather than specific angles. 

\begin{figure}[!t]
    \centering
    \includegraphics[width=\linewidth]{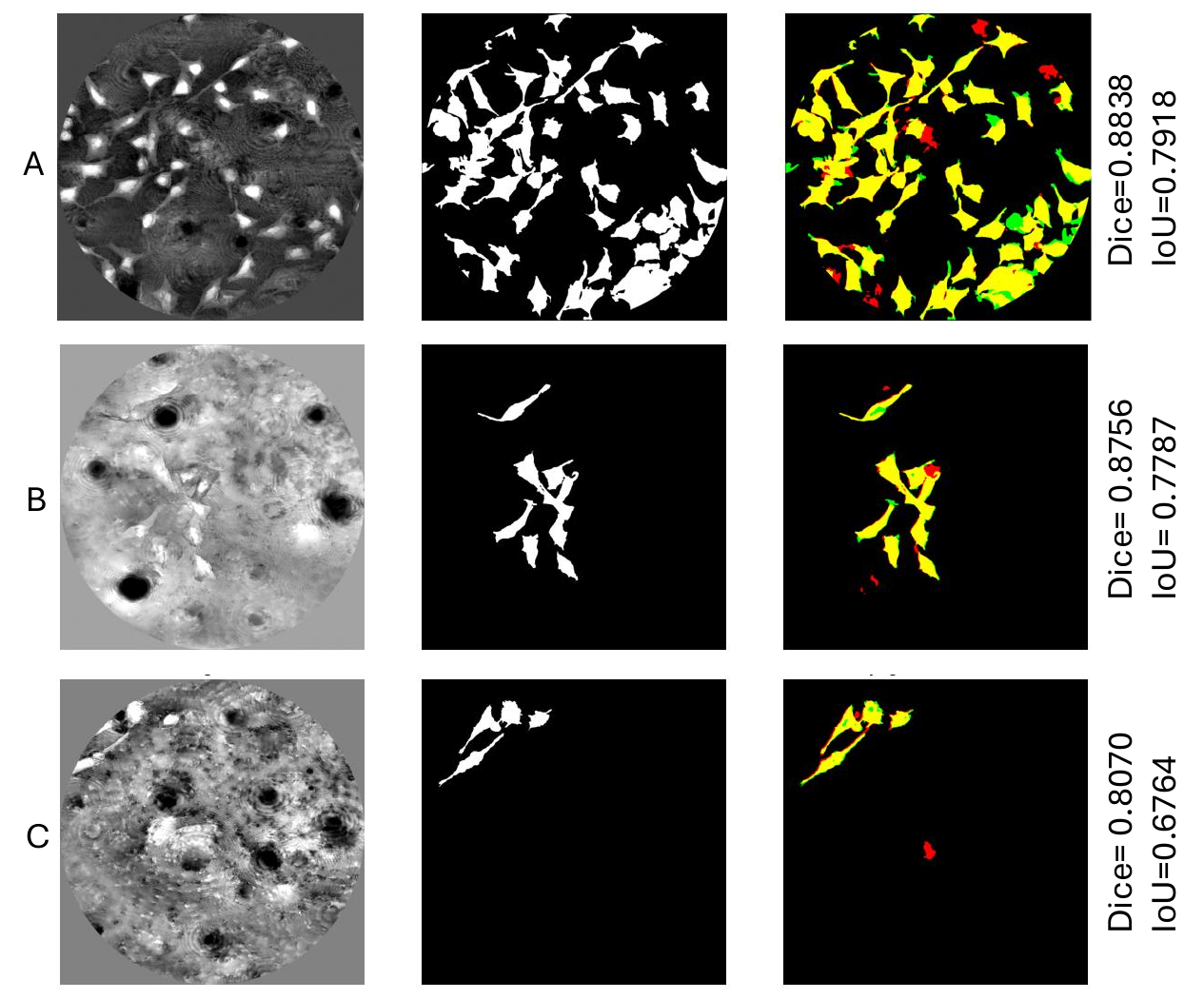}
    \caption{Qualitative segmentation on ssQPM (HeLa) at high (\textbf{A}), medium (\textbf{B}), and low (\textbf{C}) confluence. Columns show the ssQPM input, the ground-truth binary cell mask, and the DM-QPMNet prediction overlaid on the ground truth. In the overlay, \textcolor{yellow}{yellow} indicates correct agreement between prediction and ground truth, \textcolor{green}{green} shows ground-truth cell regions, and \textcolor{red}{red} shows predicted regions not supported by the ground truth. Dice/IoU for each panel are annotated.}
    \label{fig:qualitative}
\end{figure}

Figure~\ref{fig:qualitative} presents qualitative segmentation results across three density regimes.

\begin{table}[ht]
\centering
\small
\caption{Fusion ablation study.}
\label{tab:fusion_ablation}
\resizebox{\columnwidth}{!}{%
\begin{tabular}{lcc}
\hline
\textbf{Variant} & \textbf{Dice (↑)} & \textbf{IoU (↑)} \\
\hline
\rowcolor{gray!15}
\textbf{Late fusion @ Stage-2 + MHA (ours)}            & \textbf{0.888} & \textbf{0.799} \\
Late fusion @ Stage-2 \textit{w/o} MHA (concat + 1$\times$1) & 0.879 & 0.791 \\
Late fusion @ Stage-2 (cross-gating)                    & 0.875 & 0.787 \\
Early fusion (5-ch concat at input) + nnU-Net           & 0.873 & 0.785 \\
Late fusion @ Stage-3 (slightly late)                   & 0.877 & 0.789 \\
Late fusion @ Stage-1 (too early)                       & 0.871 & 0.783 \\
\hline
\end{tabular}%
}
\end{table}


\section{Conclusion}
We presented a dual-encoder nnU-Net architecture for cell segmentation in ssQPM that separates polarized intensity and phase inputs into distinct encoders before fusing them via multi-head attention. This approach enables modality-specific feature learning while leveraging complementary information from intensity maps from 4 polarization components and phase measurements captured in a single exposure.This demonstrates a promising path toward routine, label-free live-cell segmentation in ssQPM systems.

\section{Compliance with Ethical Standards}

This research was conducted retrospectively using ssQPM data made available in open access. Ethical approval was not required as confirmed by the license attached with the open-access data.

\section{Acknowledgements}
This work was supported in part by the U.S. National Science Foundation (2047592 and 2348391) and the UNC Charlotte TAIMing AI Seed Grant.

\bibliographystyle{IEEEbib}
\bibliography{refs}

\begin{thebibliography}{10}

\bibitem{Stringer2021Cellpose}
Carsen Stringer, Tim Wang, Michalis Michaelos, and Marius Pachitariu,
\newblock ``Cellpose: a generalist algorithm for cellular segmentation,''
\newblock {\em Nature Methods}, vol. 18, pp. 100--106, 2021.

\bibitem{Isensee2021nnUNet}
Fabian Isensee, Paul~F. Jaeger, Simon A.~A. Kohl, Jens Petersen, and Klaus~H. Maier-Hein,
\newblock ``nnu-net: a self-configuring method for deep learning-based biomedical image segmentation,''
\newblock {\em Nature Methods}, vol. 18, pp. 203--211, 2021.

\bibitem{Kirillov2023SAM}
Alexander Kirillov, Eric Mintun, Nikhila Ravi, et~al.,
\newblock ``Segment anything,''
\newblock {\em arXiv preprint arXiv:2304.02643}, 2023.

\bibitem{Edlund2021LIVECell}
Chris Edlund, Jens Degerman, Constantino Reyes~Aldasoro, et~al.,
\newblock ``Livecell---a large-scale dataset for label-free live cell segmentation,''
\newblock {\em Nature Methods}, vol. 18, pp. 1038--1045, 2021.

\bibitem{Ounkomol2018LabelFree}
Chawin Ounkomol, Sharmishtaa Seshamani, Mary~M Maleckar, Forrest Collman, and Gregory~R Johnson,
\newblock ``Label-free prediction of three-dimensional fluorescence images from transmitted-light microscopy,''
\newblock {\em Nature Methods}, vol. 15, no. 11, pp. 917--920, 2018.

\bibitem{Laissue2017Phototoxicity}
Philippe~P Laissue, Rana~A Alghamdi, Pavel Tomancak, Emmanuel~G Reynaud, and Hari Shroff,
\newblock ``Assessing phototoxicity in live fluorescence imaging,''
\newblock {\em Nature Methods}, vol. 14, no. 7, pp. 657--661, 2017.

\bibitem{Park2018QPIReview}
YongKeun Park, Christian Depeursinge, and Gabriel Popescu,
\newblock ``Quantitative phase imaging in biomedicine,''
\newblock {\em Nature Photonics}, vol. 12, no. 10, pp. 578--589, 2018.

\bibitem{Kim2010DHMReview}
Myung~K Kim,
\newblock ``Principles and techniques of digital holographic microscopy,''
\newblock {\em SPIE Reviews}, vol. 1, no. 1, pp. 018005, 2010.

\bibitem{Kemper2008DHM}
Bj{\"o}rn Kemper and Gert von Bally,
\newblock ``Digital holographic microscopy for live cell applications and technical inspection,''
\newblock {\em Applied Optics}, vol. 47, no. 4, pp. A52--A61, 2008.

\bibitem{Zuo2015TIE}
Chao Zuo, Qian Chen, and Anand Asundi,
\newblock ``Transport of intensity phase retrieval and computational imaging for partially coherent fields: The phase space perspective,''
\newblock {\em Optics and Lasers in Engineering}, vol. 71, pp. 20--32, 2015.

\bibitem{EspinosaMomox2024BOE}
Ana Espinosa{-}Momox, Brandon Norton, Maria Cywinska, Bryce Evans, Juan Vivero{-}Escoto, and Rosario Porras{-}Aguilar,
\newblock ``Single-shot quantitative phase microscopy: a multi-functional tool for cell analysis,''
\newblock {\em Biomedical Optics Express}, vol. 15, no. 10, pp. 5999--6012, 2024.

\bibitem{Millerd2017InstantaneousPhase}
James~E Millerd, Neal~J Brock, John~B Hayes, and James~C Wyant,
\newblock ``Instantaneous phase-shift, point-diffraction interferometer,''
\newblock {\em Proceedings of SPIE}, vol. 10380, pp. 103800D, 2017.

\bibitem{Debeir2008PhaseWatershed}
Olivier Debeir, Isabelle Adanja, Nicolas Warzee, Patrick Van~Ham, and Christine Decaestecker,
\newblock ``Phase contrast image segmentation by weak watershed transform assembly,''
\newblock in {\em IEEE International Symposium on Biomedical Imaging}, 2008, pp. 724--727.

\bibitem{Zhou2020UNetSurvey}
Zongwei Zhou, Md~Mahfuzur Rahman~Siddiquee, Nima Tajbakhsh, and Jianming Liang,
\newblock ``Unet++: Redesigning skip connections to exploit multiscale features in image segmentation,''
\newblock {\em IEEE Transactions on Medical Imaging}, vol. 39, no. 6, pp. 1856--1867, 2020.

\bibitem{Dolz2019DenseMultiPath}
Jose Dolz, Ismail~Ben Ayed, and Christian Desrosiers,
\newblock ``Dense multi-path u-net for ischemic stroke lesion segmentation in multiple image modalities,''
\newblock in {\em International MICCAI Brainlesion Workshop}. Springer, 2019, pp. 271--279.

\bibitem{Oktay2018AttentionUNet}
Ozan Oktay, Jo~Schlemper, Loic~le Folgoc, et~al.,
\newblock ``Attention u-net: Learning where to look for the pancreas,''
\newblock in {\em Proc. Medical Imaging with Deep Learning (MIDL)}, 2018.

\bibitem{Valindria2018HeteroModal}
Vanya~V. Valindria, Ioannis Lavdas, Wenjia Bai, et~al.,
\newblock ``Hetero-modal image segmentation,''
\newblock {\em Medical Image Analysis}, vol. 48, pp. 130--142, 2018.

\bibitem{Chen2021TransUNet}
Jieneng Chen, Yongyi Lu, Qihang Yu, Xiangde Luo, Ehsan Adeli, Yan Wang, Le~Lu, Alan~L Yuille, and Yuyin Zhou,
\newblock ``Transunet: Transformers make strong encoders for medical image segmentation,''
\newblock in {\em arXiv preprint arXiv:2102.04306}, 2021.

\bibitem{Valanarasu2021MedT}
Jeya Maria~Jose Valanarasu, Poojan Oza, Ilker Hacihaliloglu, and Vishal~M Patel,
\newblock ``Medical transformer: Gated axial-attention for medical image segmentation,''
\newblock in {\em International Conference on Medical Image Computing and Computer-Assisted Intervention}. Springer, 2021, pp. 36--46.

\end{thebibliography}

\end{document}